\documentclass{article}
\usepackage{ijcai19}

\usepackage{times}
\usepackage{soul}
\usepackage{url}
\usepackage[hidelinks]{hyperref}
\usepackage[utf8]{inputenc}
\usepackage[small]{caption}
\usepackage{graphicx}
\usepackage{amsmath}
\usepackage{booktabs}
\title{Pretraining-Based Natural Language Generation for Text Summarization}

\iftrue
\author{
Haoyu Zhang$^1$\footnote{Contact Author.}\and Jianjun Xu$^{1}$\and Ji Wang$^{1}$\\
\affiliations
$^1$College of Computer, National University of Defense Technology, Changsha, China\\
\emails
\{zhanghaoyu10, jjxu, wj\}@nudt.edu.cn
}
\fi

\begin{document}

\maketitle

\begin{abstract}

In this paper, we propose a novel pretraining-based encoder-decoder framework, which can generate the output sequence based on the input sequence in a two-stage manner. For the encoder of our model, we encode the input sequence into context representations using BERT.
For the decoder, there are two stages in our model, in the first stage, we use a Transformer-based decoder to generate a draft output sequence.
In the second stage, we mask each word of the draft sequence and feed it to BERT, then by combining the input sequence and the draft representation generated by BERT, we use a Transformer-based decoder to predict the refined word for each masked position.
To the best of our knowledge, our approach is the first method which applies the BERT into text generation tasks.
As the first step in this direction, we evaluate our proposed method on the text summarization task. 
Experimental results show that our model achieves new state-of-the-art on both CNN/Daily Mail and New York Times datasets.

\end{abstract}

\section{Introduction} 
Text summarization generates summaries from input documents while keeping salient information. It is an important task and can be applied to several real-world applications. Many methods have been proposed to solve the text summarization problem~\cite{See2017,NallapatiZZ17,DBLP:conf/acl/ZhaoZWYHZ18,gehrmann2018bottom}. There are two main text summarization techniques: extractive and abstractive. 
Extractive summarization generates summary by selecting salient sentences or phrases from the source text, while abstractive methods paraphrase and restructure sentences to compose the summary. We focus on abstractive summarization in this work as it is more flexible and thus can generate more diverse summaries.

Recently, many abstractive approaches are introduced based on neural sequence-to-sequence framework~\cite{Paulus2018,See2017,gehrmann2018bottom,Li2018}. 
Based on the sequence-to-sequence model with copy mechanism,~\cite{See2017} incorporates a coverage vector to track and control attention scores on source text.
~\cite{Paulus2018} introduce intra-temporal attention processes in the encoder and decoder to address the repetition and incoherent problem.

There are two issues in previous abstractive methods: 
1) these methods use left-context-only decoder, thus do not have complete context when predicting each word.
2) they do not utilize the pre-trained contextualized language models on the decoder side, so it is more difficult for the decoder to learn summary representations, context interactions and language modeling together.

Recently, BERT has been successfully used in various natural language processing tasks, such as textual entailment, name entity recognition and machine reading comprehensions. In this paper, we present a novel natural language generation model based on pre-trained language models (we use BERT in this work). As far as we know, this is the first work to extend BERT to the sequence generation task. 
To address the above issues of previous abstractive methods, in our model, we design a two-stage decoding process to make good use of BERT's context modeling ability.
On the first stage, we generate the summary using a left-context-only-decoder. 
On the second stage, we mask each word of the summary and predict the refined word one-by-one using a refine decoder. To further improve the naturalness of the generated sequence, we cooperate reinforcement objective with the refine decoder.

The main contributions of this work are:

1. We propose a natural language generation model based on BERT, making good use of the pre-trained language model in the encoder and decoder process, and the model can be trained end-to-end without handcrafted features.

2. We design a two-stage decoder process. In this architecture, our model can generate each word of the summary considering both sides' context information.

3. We conduct experiments on the benchmark datasets CNN/Daily Mail and New York Times. Our model achieves a 33.33 average of ROUGE-1, ROUGE-2 and ROUGE-L on the CNN/Daily Mail, which is state-of-the-art. On the New York Times dataset, our model achieves about 5.6\% relative improvement over ROUGE-1.

\section{Background}

\subsection{Text Summarization}

In this paper, we focus on single-document multi-sentence summarization and propose a supervised abstractive model based on the neural attentive sequence-to-sequence framework which consists of two parts: a neural network for the encoder and another network for the decoder. The encoder encodes the input sequence to intermediate representation and the decoder predicts one word at a time step given the input sequence representation vector and previous decoded output. The goal of the model is to maximize the probability of generating the correct target sequences. In the encoding and generation process, the attention mechanism is used to concentrate on the most important positions of text. The learning objective of most sequence-to-sequence models is to minimize the negative log likelihood of the generated sequence as shown in following equation, where $y^*_t$ is the t-th ground-truth summary token. 

\begin{equation}
    Loss =  - \log \sum_{t=1}^{|y|} P(y_t^*|y_{<t}^*, X)
\end{equation}

However, with this objective, traditional sequence generation models consider only one direction context in the decoding process, which could cause performance degradation since complete context of one token contains preceding and following tokens, thus feeding only preceded decoded words to the decoder so that the model may generate unnatural sequences. For example, attentive sequence-to-sequence models often generate sequences with repeated phrases which harm the naturalness. Some previous works mitigate this problem by improving the attention calculation process, but in this paper we show that feeding bi-directional context instead of left-only-context can better alleviate this problem.

\subsection{Bi-Directional Pre-Trained Context Encoders}

Recently, context encoders such as ELMo, GPT, and BERT have been widely used in many NLP tasks. These models are pre-trained on a huge unlabeled corpus and can generate better contextualized token embeddings, thus the approaches built on top of them can achieve better performance. 

Since our method is based on BERT, we illustrate the process briefly here. BERT consists of several layers. In each layer there is first a multi-head self-attention sub-layer and then a linear affine sub-layer with the residual connection. In each self-attention sub-layer the attention scores $e_{ij}$ are first calculated as Eq.~\eqref{eq:calcu_a} \eqref{eq:calcu_e}, in which $d_e$ is output dimension, and $W^Q, W^K, W^V$ are parameter matrices.

\begin{eqnarray}
    &a_{ij} = \cfrac {(h_iW_Q)(h_jW_K)^T}  {\sqrt{d_e}} \label{eq:calcu_a} \\
    &e_{ij} = \cfrac {\exp{e_{ij}}} {\sum_{k=1}^N\exp{e_{ik}}} \label{eq:calcu_e}
\end{eqnarray}

Then the output is calculated as Eq.~\eqref{eq:calcu_o}, which is the weighted sum of previous outputs $h$ added by previous output $h_i$. The last layer outputs is context encoding of input sequence.

\begin{equation}
    o_i = h_i + \sum_{j=1}^{N} e_{ij}(h_j W_V) \label{eq:calcu_o}
\end{equation}

Despite the wide usage and huge success, there is also a mismatch problem between these pre-trained context encoders and sequence-to-sequence models. The issue is that while using a pre-trained context encoder like BERT, they model token-level representations by conditioning on both direction context. During pre-training, they are fed with complete sequences. However, with a left-context-only decoder, these pre-trained language models will suffer from incomplete and inconsistent context and thus cannot generate good enough context-aware word representations, especially during the inference process. 

\section{Model}

In this section, we describe the structure of our model, which learns to generate an abstractive multi-sentence summary from a given source document. 

Based on the sequence-to-sequence framework built on top of BERT, we first design a refine decoder at word-level to tackle the two problems described in the above section. We also introduce a discrete objective for the refine decoders to reduce the exposure bias problem. The overall structure of our model is illustrated in Figure~\ref{fig:model}. 

\subsection{Problem Formulation}

We denote the input document as $X = \{x_1, \ldots, x_m\}$ where $x_i \in \mathcal{X}$ represents one source token. The corresponding summary is denoted as $Y = \{y_1, \ldots, y_L\}$, $L$ represents the summary length.

 Given input document $X$, we first predict the summary draft by a left-context-only decoder, and then using the generated summary draft we can condition on both context sides and refine the content of the summary. The draft will guide and constrain the refine process of summary.
 
\subsection{Summary Draft Generation}

The summary draft is based on the sequence-to-sequence model. On the encoder side the input document $X$ is encoded into representation vectors $H = \{h_1, \ldots, h_m\}$, and then fed to the decoder to generate the summary draft $A = \{a_1, \ldots, a_{|a|}\}$.

\subsubsection{Encoder}

We simply use BERT as the encoder. It first maps the input sequence to word embeddings and then computes document embeddings as the encoder's output, denoted by following equation.

\begin{equation}
    H = BERT(x_1, \ldots, x_m)
\end{equation}

\begin{figure*}[htbp]
	\centering
	\small
	\includegraphics[width=0.98\textwidth]{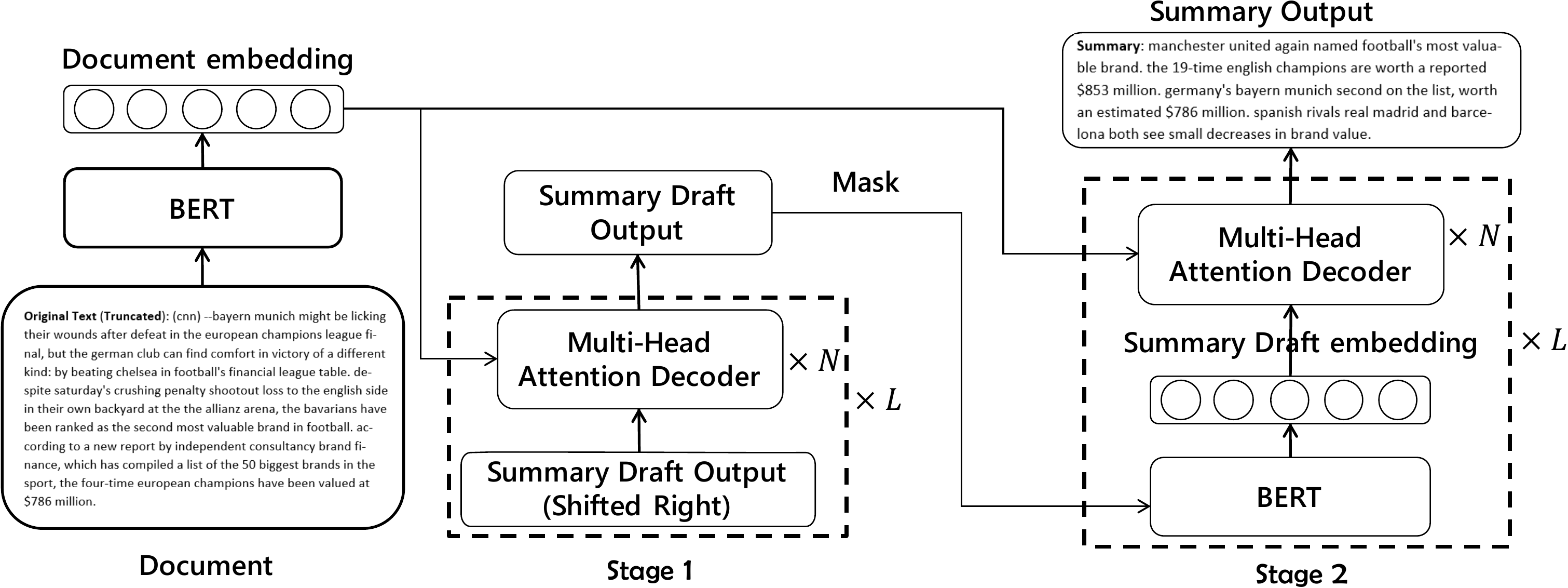}
	\makeatletter\def\@captype{figure}\makeatother \caption{Model Overview, N represents decoder layer number and L represents summary length.\label{fig:model}}
\end{figure*}

\subsubsection{Summary Draft Decoder}

In the draft decoder, we first introduce BERT's word embedding matrix to map the previous summary draft outputs $\{y_1, \ldots, y_{t-1}\}$ into embeddings vectors $\{q_1, \ldots, q_{t-1}\}$ at t-th time step. Note that as the input sequence of the decoder is not complete, we do not use the BERT network to predict the context vectors here.

Then we introduce an $N$ layer Transformer decoder to learn the conditional probability $P(A|H)$. Transformer's encoder-decoder multi-head attention helps the decoder learn soft alignments between summary and source document. At the t-th time step, the draft decoder predicts output probability conditioned on previous outputs and encoder hidden representations as shown in Eq.~\eqref{eq:p_vocab}, in which $q_{<t} = \{q_1, \ldots, q_{t-1}\}$. Each generated sequence will be truncated in the first position of a special token '[PAD]'. The total summary draft decoder progress is shown in Stage 1 of Figure~\ref{fig:model}.

\begin{eqnarray}
    &P^{vocab}_t(w) = f_{dec}(q_{<t}, H) \label{eq:p_vocab} \\
    &L_{dec} = \sum_{t=1}^{|a|} -\log P(a_t = y_t^*|a_{< t}, H) \label{eq:obj_dec}
\end{eqnarray}

As Eq.~\eqref{eq:obj_dec} shows, the decoder's learning objective is to minimize negative likelihood of conditional probability, in which $y_t^*$ is the t-th ground truth word of summary. 

However a decoder with this structure is not sufficient enough: if we use the BERT network in this decoder, then during training and inference, in-complete context(part of sentence) is fed into the BERT module, and although we can fine-tune BERT's parameters, the input distribution is quite different from the pre-train process, and thus harms the quality of generated context representations.

If we just use the embedding matrix here, 
it will be more difficult for the decoder with fresh parameters to learn to model representations as well as vocabulary probabilities, from a relative small corpus compared to BERT's huge pre-training corpus. In a word, the decoder cannot utilize BERT's ability to generate high quality context vectors, which will also harm performance.

This issue exists when using any other contextualized word representations, so we design a refine process to mitigate it in our approach which will be described in the next sub-section.

\subsubsection{Copy Mechanism}

As some summary tokens are out-of-vocabulary words and occurs in input document, we incorporate copy mechanism~\cite{Gu2015} based on the Transformer decoder, we will describe it briefly.

At decoder time step $t$, we first calculate the attention probability distribution over source document $X$ using the bi-linear dot product of the last layer decoder output of Transformer $o_t$ and the encoder output $h_j$, shown in Eq.~\eqref{eq:et} \eqref{eq:at}. 

\begin{eqnarray}
u_t^j =& o_t W_c h_j \label{eq:et}\\
\alpha_t^j =& \cfrac {\exp{u_t^j}}   {\sum_{k=1}^N\exp{u_t^k}} \label{eq:at} 
\end{eqnarray}

We then calculate copying gate $g_t\in [0, 1]$, which makes a soft choice between selecting from source and generating from vocabulary, $W_c, W_g, b_g$ are parameters: 

\begin{equation}
g_t = sigmoid(W_g \cdot [o_t, h] + b_g) \label{eq:copy_prob}
\end{equation}

Using $g_t$ we calculate the weighted sum of copy probability and generation probability to get the final predicted probability of extended vocabulary $\mathcal{V} + \mathcal{X}$, where $\mathcal{X}$ is the set of out of vocabulary words from the source document. The final probability is calculated as follow:

\begin{equation}
P_t(w) = (1-g_t)P_t^{vocab}(w) + g_t\sum_{i:w_i=w} \alpha_t^i
\end{equation}

\subsection{Summary Refine Process}

The main reason to introduce the refine process is to enhance the decoder using BERT's contextualized representations, so we do not modify the encoder and reuse it during this process.

On the decoder side, we propose a new word-level refine decoder. The refine decoder receives a generated summary draft as input, and outputs a refined summary. As Figure~\ref{fig:model} Stage 2 shows, it first masks each word in the summary draft one by one, then feeds the draft to BERT to generate context vectors. Finally it predicts a refined summary word using an $N$ layer Transformer decoder which is the same as the draft decoder. At t-th time step the t-th word of input summary is masked, and the decoder predicts the refined word given other words of the summary.

The learning objective of this process is shown in Eq.~\eqref{eq:obj_word}, $y_t$ is the t-th summary word and $y_{t}^*$ for the ground-truth summary word, and $a_{\neq t} = \{a_1, \ldots, a_{t-1}, a_{t+1}, \ldots, a_{|y|}\}$.

\begin{equation}
    L_{refine} = \sum_{t=1}^{|y|} -\log P(y_t = y_t^*|a_{\neq t}, H) \label{eq:obj_word}
\end{equation}

From the view of BERT or other contextualized embeddings, the refine decoding process provides a more complete input sequence which is consistent with their pre-training processes. Intuitively, this process works as follows: first the draft decoder writes a summary draft based on a document, and then the refine decoder edits the draft. It concentrates on one word at a time, based on the source document as well as other words.

We design the word-level refine decoder because this process is similar to the cloze task in BERT's pre-train process, therefore by using the ability of the contextual language model the decoder can generate more fluent and natural sequences.

The parameters are shared between the draft decoder and refine decoder, as we find that using individual parameters the model's performance degrades a lot. The reason may be that we use teach-forcing during training, and thus the word-level refine decoder learns to predict words given all the other ground-truth words of summary. This objective is similar to the language model's pre-train objective, and is probably not enough for the decoder to learn to generate refined summaries. So in our model all decoders share the same parameters.

\subsubsection{Mixed Objective}

For summarization, ROUGE is usually used as the evaluation metric, however during model training the objective is to maximize the log likelihood of generated sequences. This mis-match harms the model's performance. Similar to previous work~\cite{kryscinski2018improving}, we add a discrete objective to the model, and optimize it by introducing the policy gradient method. The discrete objective for the summary draft process is as shown in Eq.~\eqref{eq: rl}, where $a^s$ is the draft summary sampled from predicted distribution, and $R(a^s)$ is the reward score compared with the ground-truth summary, we use ROUGE-L in our experiment. To balance between optimizing the discrete objective and generating readable sequences, we mix the discrete objective with maximum-likelihood objective. Eq.~\eqref{eq: mixed} shows the final objective for the draft process, note here $L_{dec}$ is $-logP(a|x)$. In the refine process we introduce similar objectives. 

\begin{eqnarray}
    L^{rl}_{dec} = R(a^s)\cdot  [-\log (P(a^s|x))] \label{eq: rl} \\
    \hat L_{dec} = \gamma * L^{rl}_{dec} + (1 - \gamma) * L_{dec} \label{eq: mixed}
\end{eqnarray}

\subsection{Learning and Inference}

During model training, the objective of our model is sum of the two processes, jointly trained using "teacher-forcing" algorithm. During training we feed the ground-truth summary to each decoder and minimize the following objective.

\begin{equation}
    L_{model} = \hat L_{dec} + \hat L_{refine}
\end{equation}

At test time, each time step we choose the predicted word by $\hat y = argmax_{y'} P(y'|x)$, use beam search to generate the draft summaries, and use greedy search to generate the refined summaries.

\section{Experiment}

\begin{table*}[tpb]  
\centering
\begin{tabular}{l|lll|l}
\hline
Model                 & ROUGE-1                     & ROUGE-2       &      ROUGE-L    &  R-AVG     \\ \hline
Extractive \\ \hline
lead-3~\cite{See2017}   & 40.34    &  17.70   &      36.57     &   31.54        \\
SummmaRuNNer~\cite{NallapatiZZ17}     & 39.60     &  16.20         &      35.30     &   30.37         \\ 
Refresh~\cite{Narayan2018}               & 40.00                       &  18.20         &      36.60     & 31.60            \\ 
DeepChannel~\cite{Shi2018}  & 41.50                       &  17.77         &      37.62   & 32.30        \\
rnn-ext + RL~\cite{chen2018fast}          & 41.47                       &  18.72         &      37.76   & 32.65            \\
MASK-$LM^{global}$~\cite{Chang2019}   & 41.60 & 19.10 & 37.60 & 32.77 \\
NeuSUM~\cite{DBLP:conf/acl/ZhaoZWYHZ18}              & 41.59                       &  19.01         &      37.98     & 32.86          \\
\hline
Abstractive \\ \hline
PointerGenerator+Coverage~\cite{See2017}      & 39.53                       & 17.28         &      36.38      & 31.06           \\
ML+RL+intra-attn~\cite{Paulus2018} & 39.87                  &  15.82        &      36.90      & 30.87           \\ 
inconsistency loss\cite{hsu2018unified}    & 40.68                       &  17.97        &      37.13      & 31.93      \\
Bottom-Up Summarization~\cite{gehrmann2018bottom} & 41.22                     &  18.68        &      38.34      & 32.75      \\
DCA~\cite{Celikyilmaz2018}                   & 41.69                       &  19.47        &      37.92      & 33.11         \\ \hline
Ours \\ \hline
One-Stage           &39.50        & 17.87          & 36.65  & 31.34  \\   
Two-Stage         & 41.38  & 19.34 & 38.37       & 33.03 \\
Two-Stage + RL               & \textbf{41.71} & \textbf{19.49} & \textbf{38.79}               & \textbf{33.33}                      \\ \hline
\end{tabular}
\caption{ROUGE F1 results for various models and ablations on the CNN/Daily Mail test set. R-AVG calculates average score of Rouge-1, Rouge-2 and Rouge-L.\label{r1}}
\end{table*}

\subsection{Settings}

In this work, all of our models are built on $BERT_{BASE}$, although another larger pre-trained model with better performance ($BERT_{LARGE}$) has published but it costs too much time and GPU memory. We use WordPiece embeddings with a 30,000 vocabulary which is the same as BERT. We set the layer of transformer decoders to 12(8 on NYT50), and set the attention heads number to 12(8 on NYT50), set fully-connected sub-layer hidden size to 3072. We train the model using an Adam optimizer with learning rate of $3e-4$, $\beta_1=0.9$, $\beta_2=0.999$ and $\epsilon=10^{-9}$ and use a dynamic learning rate during the training process. For regularization, we use dropout~\cite{Srivastava2014Dropout} and label smoothing~\cite{Szegedy2015Rethinking} in our models and set the dropout rate to 0.15, and the label smoothing value to 0.1. We set the RL objective factor $\gamma$ to 0.99.

During training, we set the batch size to 36, and train for 4 epochs(8 epochs for NYT50 since it has many fewer training samples), after training the best model are selected from last 10 models based on development set performance. Due to GPU memory limit, we use gradient accumulation, set accumulate step to 12 and feed 3 samples at each step. We use beam size 4 and length penalty of 1.0 to generate logical form sequences.

We filter repeated tri-grams in beam-search process by setting word probability to zero if it will generate an tri-gram which exists in the existing summary. It is a nice method to avoid phrase repetition since the two datasets seldom contains repeated tri-grams in one summary. We also fine tune the generated sequences using another two simple rules. When there are multi summary sentences with exactly the same content, we keep the first one and remove the other sentences; we also remove sentences with less than 3 words from the result.

\subsubsection{Datasets}
To evaluate the performance of our model, we conduct experiments on CNN/Daily Mail dataset, which is a large collection of news articles and modified for summarization. Following \cite{See2017} we choose the non-anonymized version of the dataset, which consists of more than 280,000 training samples and 11490 test set samples. 

We also conduct experiments on the New York Times(NYT) dataset which also consists of many news articles. The original dataset can be applied here.\footnote{http://duc.nist.gov/} In our experiment, we follow the dataset splits and other pre-process settings of ~\cite{Durrett2016}.  We first filter all samples without a full article text or abstract and then remove all samples with summaries shorter than 50 words. Then we choose the test set based on the date of publication(all examples published after January 1, 2007). The final dataset contains 22,000 training samples and 3,452 test samples and is called NYT50 since all summaries are longer than 50 words.

We tokenize all sequences of the two datasets using the WordPiece tokenizer. After tokenizing, the average article length and summary length of CNN/Daily Mail are 691 and 51, and NYT50's average article length and summary length are 1152 and 75. We truncate the article length to 512, and the summary length to 100 in our experiment(max summary length is set to 150 on NYT50 as its average golden summary length is longer).

\subsubsection{Evaluation Metrics}
On CNN/Daily Mail dataset, we report the full-length F-1 score of the ROUGE-1, ROUGE-2 and ROUGE-L metrics, calculated using PyRouge package\footnote{pypi.python.org/pypi/pyrouge/0.1.3} and the Porter stemmer option. On NYT50, following \cite{Paulus2018} we evaluate limited length ROUGE recall score(limit the generated summary length to the ground truth length). We split NYT50 summaries into sentences by semicolons to calculate the ROUGE scores.

\subsection{Results and Analysis}

Table \ref{r1} gives the results on CNN/Daily Mail dataset, we compare the performance of many recent approaches with our model. We classify them to two groups based on whether they are extractive or abstractive models. As the last line of the table lists, the ROUGE-1 and ROUGE-2 score of our full model is comparable with DCA, and outperforms on ROUGE-L. Also, compared to extractive models NeuSUM and MASK-$LM^{global}$, we achieve slight higher ROUGE-1.
Except the four scores, our model outperforms these models on all the other scores, and since we have 95\% confidence interval of at most $\pm$ 0.20, these improvements are statistically significant.

\subsubsection{Ablation Analysis}

As the last four lines of Table~\ref{r1} show, we conduct an ablation study on our model variants to analyze the importance of each component. 
We use three ablation models for the experiments. 
One-Stage: A sequence-to-sequence model with copy mechanism based on BERT; 
Two-Stage: Adding the refine decoder to the One-Stage model; 
Two-Stage + RL: Full model with refine process cooperated with RL objective.

First, we compare the Two-Stage+RL model with Two-Stage ablation, we observe that the full model outperforms by 0.30 on average ROUGE, suggesting that the reinforcement objective helps the model effectively. 
Then we analyze the effect of refine process by removing it from the Two-Stage model, we observe that without the refine process the average ROUGE score drops by 1.69. 
The ablation study proves that each module is necessary for our full model, and the improvements are statistically significant on all metrics.

\begin{figure}[htbp]
	\centering
	\includegraphics[width=0.5\textwidth]{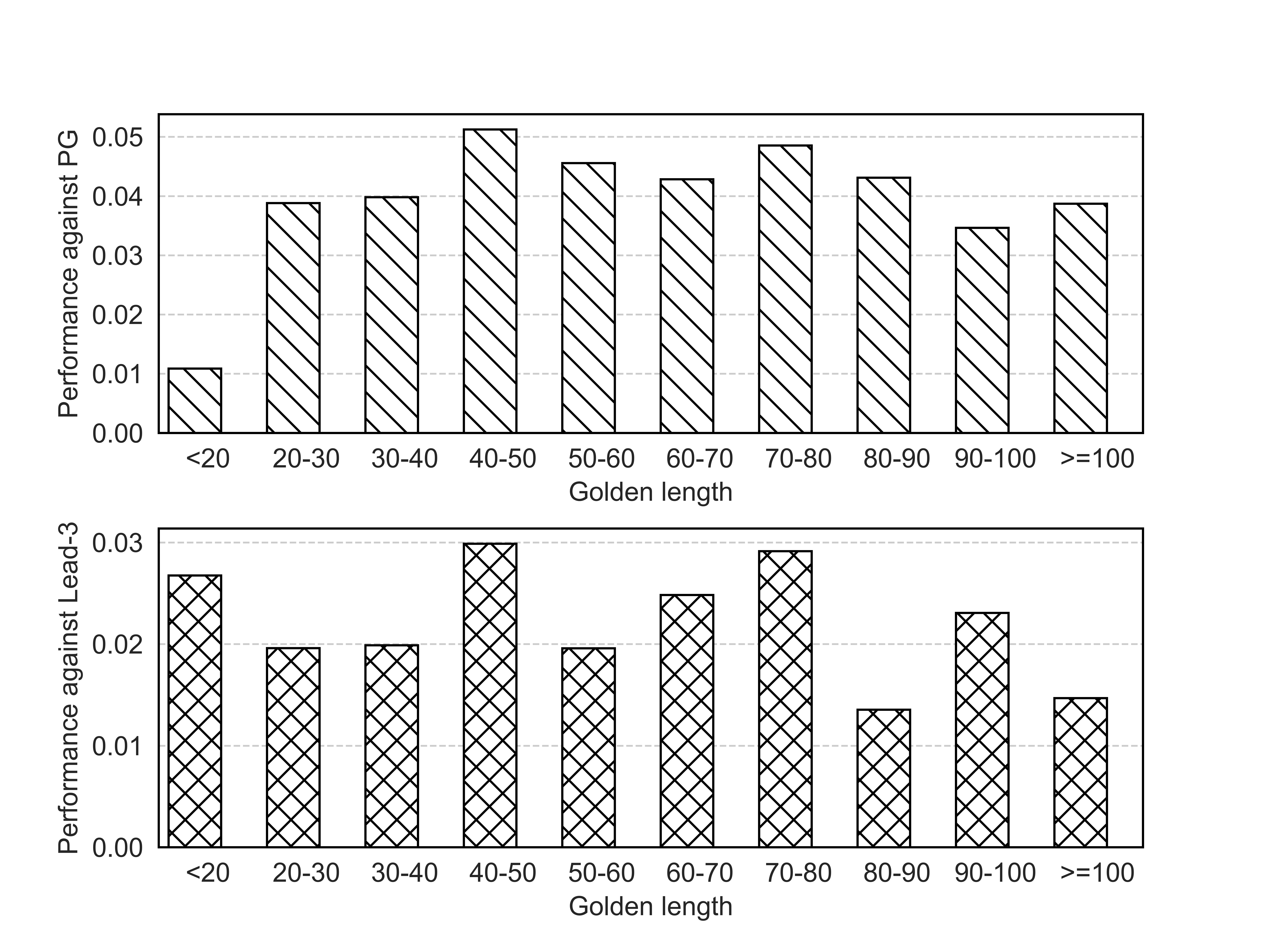}
	\makeatletter\def\@captype{figure}\makeatother \caption{Average ROUGE-L improvement on CNN/Daily mail test set samples with different golden summary length.\label{fig:result}}
\end{figure}

\subsubsection{Effects of Summary Length}

To evaluate the impact of summary length on model performance, we compare the average rouge score improvements of our model with different length of ground-truth summaries. 
As the above sub-figure of Figure~\ref{fig:result} shows, compared to Pointer-Generator with Coverage, on length interval 40-80(occupies about 70\% of test set) the improvements of our model are higher than shorter samples, confirms that with better context representations, in longer documents our model can achieve higher performance.

As shown in the below sub-figure of Figure~\ref{fig:result}, compared to extractive baseline: Lead-3~\cite{See2017}, the advantage of our model will fall when golden summary length is greater than 80. This probably because that we truncate the long documents and golden summaries and cannot get full information, it could also because that the training data in these intervals is too few to train an abstractive model, so simple extractive method will not fall too far behind.

\begin{table}[htpb]  
\centering
\begin{tabular}{l|ll}
\hline
Model                 & R-1                         & R-2       \\ \hline
First sentences	      & 28.6         		        &  17.3           \\
First $k$ words  & 35.7         		        &  21.6              \\ \hline
Full~\cite{Durrett2016}      & 42.2                        &  24.9      \\ \hline
ML+RL+intra-attn~\cite{Paulus2018} & 42.94                 &  26.02  \\ \hline
Two-Stage + RL~(Ours)                       & \textbf{45.33} & \textbf{26.53}      \\ \hline
\end{tabular}
\caption{Limited length ROUGE recall results on the NYT50 test set.\label{r2}}
\end{table}

\subsection{Additional Results on NYT50}

Table~\ref{r2} reports experiment results on the NYT50 corpus. Since the short summary samples are filtered, NYT50 has average longer summaries than CNN/Daily Mail. So the model needs to catch long-term dependency of the sequences to generate good summaries.

The first two lines of Table~\ref{r2} show results of the two baselines introduced by~\cite{Durrett2016}: these baselines select first n sentences, or select the first k words from the original document. Also we compare performance of our model with two recent models, we see 2.39 ROUGE-1 improvements compared to the ML+RL with intra-attn approach(previous SOTA) carries over to this dataset, which is a large margin. On ROUGE-2, our model also get an improvement of 0.51. The experiment proves that our approach can outperform competitive methods on different data distributions.

\section{Related work}

\subsection{Text Summarization}

Text summarization models are usually classified to abstractive and extractive ones.
Recently, extractive models like DeepChannel~\cite{Shi2018}, rnn-ext+RL~\cite{chen2018fast} and NeuSUM~\cite{DBLP:conf/acl/ZhaoZWYHZ18} achieve higher performances using well-designed structures. For example, DeepChannel propose a salience estimation network and iteratively extract salient sentences.~\cite{zhang2018neural} train a sentence compression model to teach another latent variable extractive model.

Also, several recent works focus on improving abstractive methods.~\cite{gehrmann2018bottom} design a content selector to over-determine phrases in a source document that should be part of the summary. ~\cite{hsu2018unified} introduce inconsistency loss to force words in less attended sentences(which determined by extractive model) to have lower generation probabilities.
~\cite{Li2018} extend seq2seq model with an information selection network to generate more informative summaries. 

\subsection{Pre-trained language models}

Pre-trained word vectors~\cite{mikolov2013distributed,pennington2014glove,bojanowski2017enriching} have been widely used in many NLP tasks. More recently, pre-trained language models (ELMo, GPT and BERT), have also achieved great success on several NLP problems such as textual entailment, semantic similarity, reading comprehension, and question answering ~\cite{peters2018deep,radford2018improving,devlin2018bert}. 

Some recent works also focus on leveraging pre-trained language models in summarization.~\cite{Radford2017} pretrain a language model and use it as the sentiment analyser when generating reviews of goods.
~\cite{kryscinski2018improving} train a language model on golden summaries, and then use it on the decoder side to incorporate prior knowledge. 

In this work, we use BERT(which is a pre-trained language model using large scale unlabeled data) on the encoder and decoder of a seq2seq model, and by designing a two stage decoding structure we build a competitive model for abstractive text summarization.

\section{Conclusion and Future Work}
In this work, we propose a two-stage model based on sequence-to-sequence paradigm. Our model utilize BERT on both encoder and decoder sides, and introduce reinforce objective in learning process. We evaluate our model on two benchmark datasets CNN/Daily Mail and New York Times, the experimental results show that compared to previous systems our approach effectively improves performance. 

Although our experiments are conducted on summarization task, our model can be used in most natural language generation tasks, such as machine translation, question generation and paraphrasing.  The refine decoder and mixed objective can also be applied on other sequence generation tasks, and we will investigate on them in future work.

\bibliographystyle{named}
\bibliography{ijcai19}

\end{document}